\documentclass[conference]{IEEEtran}
\usepackage{cite}
\usepackage{amsmath}
\usepackage{graphicx}
\usepackage{hyperref}
\usepackage{subfigure}
\begin{document}
\title{Detecting Urban Dynamics Using Deep Siamese Convolutional Neural Networks}
\author{\IEEEauthorblockN{Ephrem Admasu Yekun}
\IEEEauthorblockA{Ethiopian Institute of Technology - Mekelle\\
School of Electrical and Computer Engineering\\
Email: ephrem.admasu@mu.edu.et}
\and
\IEEEauthorblockN{Petros Reda Samsom}
\IEEEauthorblockA{Institute of Geo-Information and Earth
Observation Science\\
Department of Geo-Informatics\\
Email: Petros.r@gmail.com}}

\maketitle

\begin{abstract}
Change detection is a fast growing descipline in the areas of computer vison and remote sensing. In this work, we designed and developed a
variant of convolutional neural network (CNN), known as Siamese CNN to extract features from pairs of Sentinel-2 temporal images of Mekelle city captured at different times
and detect changes due to urbanization: buildings
and roads. The detection capability of the proposed was measured in terms of overall
accuracy (95.8\%), Kappa measure (72.5\%), recall (76.5\%), precision (77.7\%), F1 measure (77.1\%). The model has achieved a good performance in terms of most of these measures
and can be used to detect changes in Mekelle and other cities at different time horizons
undergoing urbanization.
\end{abstract}


\section{Introduction}
The population of the earth is in constant growth and more people are moving to urban areas. This puts a lot of strain on big cities and their infrastructure. Homes to live in, office quarters, industrial buildings, public service facilities, roads, and other transportation means need to be fitted in the already built-up cities to house the life of the newcomers \cite{smits1999updating}. The same happens in all cities of Ethiopia especially the capital cities of regional states.  In Mekelle, urban areas are rapidly changing mainly due to human activities in construction, destruction of topographic elements such as buildings and roads. These changes in urban dynamics enforce updating of old records, which can help planners to have accurate building zones, road infrastructure for urban planning, maintenance, and development. Thus, remote sensing increasingly becomes an essential field for up-to-date, area-wide, and cost-effective data acquisition, especially in explosively expansive cities of developing countries like Ethiopia.

The extraction of tangible and reliable information from remote sensing data places an essential role in the conservation of natural resources, environmental monitoring, spatial planning enforcement, and habitat preservation. With the help of satellite data generators such as Landsat and SPOT satellites or the ASTER and MODIS instruments and commercial satellites such as WorldView-3 (0.31m resolution panchromatic), it is possible to study and control biodiversity, food security, deforestation, and desertification, and detect vehicles and provide security services.

Urban area detection from very high resolution (VHR) remote sensing images has been widely used in applications such as urban development analysis, military reconnaissance, map updating, and disaster management \cite{zheng2014urban, shi2015accurate, ma2015robust, li2016unsupervised}. The detection procedure in those application areas was done manually and hence it was time-consuming processing since the images are high resolution and cover large areas. Nowadays, however, we are witnessing automated real-time urban area detection for VHR remote sensing images \cite{tian2018urban}. Although the study of urban expansion and its impacts is well-founded for large cities, the shortage of up to date and reliable demographic and spatial data left the study for small- and mid-cities in its immature stage. Meanwhile, the advancement of remote sensing technologies with cheap and good quality satellite images and the geographical information system (GIS) has led researchers to conduct qualitative analysis urban expansion with improved accuracy and lesser cost \cite{epstein2002techniques}.

This research focuses on the estimation of changed areas due to urbanization of Mekelle City using satellite imagery to extract features using deep convolutional neural networks (DCNN), and identify the Spatio-temporal expansion of the identified features dating back to 2015. The model is developed to identify features from very high-resolution image satellite derivatives for feature extraction. For this thesis we developed an approach of deep learning feature extraction which focuses on the following major methods: 
\begin{enumerate}
\itemsep0em
\item Feature extraction, 
\item Classification 
\item Change detection and
\item Performance assessment
\end{enumerate}
Change detection of remote sensing areas can be formally defined as follows: given two high-resolution remote sensing images $I_{t_1}$ and $I_{t_2}$, captured at different times $t_1$ and $t_2$ $, t_2 > t_1$, the change detection algorithm generates a binary image, $I_B$, which identifies change and unchanged regions due to urbanization and the changes are identified using the rule:

\[ I_B(x) = \left\{ \begin{array}{ll}
1 & \mbox{if change is detected at pixel x},\\
0 & \mbox{otherwise}
\end{array}\right.\]

Several studies have been proposed to generate the change map of Mekelle city and none of these use deep learning techniques such as CNN. With enough dataset CNNs have proven to most accurate results for change detection. Therefore, our work mainly focuses on using a variant of CNN, known as Siamese CNN, to generate the change map of Mekelle city as changed and unchaged pixels due to urbanization. To the best of our knowledge, this is the first time a Siamese CNN is used for change detection generation of Mekelle city

\section{Literature review}

\subsection{Change detection methods}
The science of change detection has been around for a long time. Early change detection methods used field survey to manage and process individual images which was time consuming and error prone \cite{singh1989review, hussain2013change}. With programs such as Copernicus and Landsat, it is now possible to access large amounts of earth observation imagery. Research on change detection is an active area and new methods are still being introduced. 

\begin{figure}[h]
\centering
\includegraphics[scale=.46]{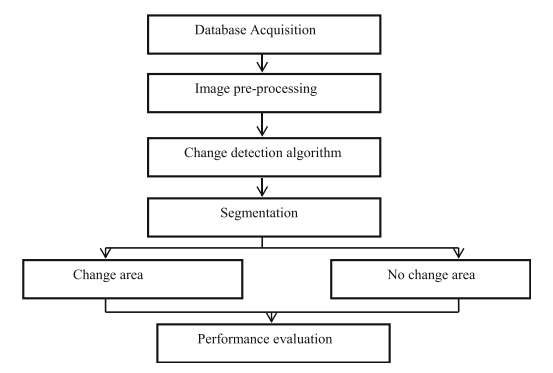}

\caption{\label{cdmethods} Scheme for change detection \cite{asokan2019change}}
\end{figure}

While implementing a change detection task, important steps have to be considered (See Figure \ref{cdmethods}). The first step in implementing change detection is image acquisition--obtaining images from satellites. The images captured from satellites differ in terms of brightness, wavelength, color and resolution. The resolution of an image represents how much information stored in each pixel. The altitude of the satellite orbit and the sensor used are determinat factors of the resolution of the image. Well-known categories of satellite images include: VISible images (VIS), InfraRed images (IR), Multispectral images, Hyperspectral images, and Synthetic Aperture Radar (SAR).

The ultimate goal of any change detection method is to identify changed pixels of the same satellite images captured at different times and ignore changes caused by atmospheric conditions (such as clouds), illumination changes, and sensor viewing angles. We can categorize change detection metods into seven: \textit{algebra, transformation, classification, advanced models, Geographical Information System (GIS)}, and \textit{other} approaches. 

Algebra based methods use simple algebraic operations to determine the changed pixels in an image. Most commonly used types of algebra-based methods include image differencing, image ratioing, vegetation index differencing, and change vector analysis (CVA). In image differencing, the difference between each pixel in both image is calcualted and changes are detected from the difference matrix. Image ratioing \cite{lee2004detecting} computes the ratio, instead of the difference, between the pixels the two images. Change vector analysis (CVA) \cite{bruzzone2002adaptive, bruzzone2000automatic, malila1980change} generates a feature vector for every pixel of the images and the magnitude of the difference between the two feature vectors is taken to describe the change. CVAs are mainly used for multispectral images. Researchers have proposed different criterias to choose the threshold value \cite{rosin1998thresholding, rosin2003evaluation} where a detailed survey and experimental setup was used on several scenarios for selecting the threshold value.

Transform based methods detect changes in images by first transforming the pixels into another dimension. Well-known techniques include Principal Component Analysis (PCA) \cite{sadeghi2016design}, Tasselled Cap Transform \cite{massarelli2018fast, thakkar2016effective, solano2018approach}, Gram-Schmidt Transformation \cite{collins1994change}, and Chi-Square Transform \cite{vazquez2017applying}. Transform based techniques have a standard pipeline of operation. First the redundant information in bands is reduced by decomposing the image into a set of independent components; the independent components are analyzed and separated and transformed into image components. Finally, a change map is produced based on some threshold value \cite{asokan2019change}. Transform based methods are effective for reducing reducing redundancy between bands and providing important information in the derived images. Nevertheless,
the change information in the trnasformed image can be difficult to interpret and label and identifying changes requires careful selection of thresholds.

Classification based methods includes post-classification comparison \cite{radhika2018neural}, Expectation Maximization (EM) algoirthms \cite{prendes2014new}, unsupervised methods \cite{vignesh2016novel}, and Artificial Neural Network (ANN) methods \cite{radhika2018neural}. Classification based CD methods require a large high quality training images to obtain good accuracy of CD and they are unaffected by external factors such as atmospheric artefacts. The perfomrance of these techniques is compromised, however, when high quality data set of large size is scarce. Researchers have used different classificaion methods for CD. Classification based method have become a popular choice for change detection. However, the acquisition of training samples for classification is difficult and time consuming and shortage of quality data can result in poor performance of overall CD model. 

The advanced change detection models include the Li--Strahler
reflectance model \cite{zeng2008scaling}, spectral mixture models \cite{xu2017detection}, and biophysical parameter estimation models \cite{yan2018novel}. The methods use linear or nonlinear models to convert the reflectance values of images to physically interpretable parameters which are easier to interpret and extract vegetation information than spectral parameters. However, these methods are computationally expensive and the models can be complicated to develop. In general advaced model based CD is effective for accurately classifying changes; however, it is difficult to develop models for reflectance value conversion.

GIS based methods incorporate varieties of source data into the change detection applications. These methods give a considerable view and coverage of the area under study. But the different source data causes the change detection method to have lower accuracy. In \cite{maurya2016evaluation} remote sensing and GIS were used to evaluate the characteristic change as part of the historical changes of the river course in India. The GIS inclusion helped the integration of source information easily and visualization of changes appropriately. A vegetation cover change detection based on Normalized Difference Vegetation Index (NDVI) using GIS information and remote sensing images was employed in \cite{gandhi2015ndvi}. In \cite{rawat2015monitoring}  discussed the spatio-temporal dynamics of land cover land use  of Hawalbagh block of district Almora, Uttarakhand, India using remote sensing and GIS which contributed in making change detection procedure easier and boosted the accuracy. 

The methods are CD techniques that do not belong to either of the five methods mentioned above and whose practical use has not been well-researched. In \cite{feng2018novel}, object-based urban CD was implemented by first distinguishing consistent objects based on vector-raster integration and image segmentation and then using Rotation Forest and coarse-to-fine uncertainty analyses. Multivariate Adaptive Regression Splines (MARS) model and Back-Propagation Neural Networks (BPNNs) were used in \cite{yang2017robust} for vegetation cover change analysis. The main advantage of the proposed model is that it  applies data augmentation techniques: flipping, translation, and rotation to deal with the scarcity of training images and produce competetive accuracy results.

\subsection{Recent related works}

Due to the lack of large training datasets, recent deep learning methods use transfer learning for change detection, which is a technique of taking a network trained on huge dataset of different problem and apply to a different but related problem \cite{el2017zoom, liu2016deep, gong2015change}. The use of transfer learning is a valid method but comes with restricting feature; the big CNNs used for transfer learning are trained on RGB images, while most remote sensing images used for change detection have several useful bands (e.g. Sentinel-2 contains 13 useful bands). Another way of using supervised deep learning methods for change detection is to generate a difference image from the trained model and manually threshold the change map \cite{zhan2017change}. However, such methods fail to incorporate end-to-end training, which can produce better results when trained successfully \cite{daudt2018urban}.

Change detection using Siamese and Early fusion architectures was in used by \cite{daudt2018urban} for change detection. These architectures were trained on a new freely available dataset, The Onera Satellite
Change Detection (OSCD) dataset, which is composed of pairs of multispectral aearial images with changes manually annotated at pixel. The method achieved higher performance than other competetive methods; however, it has not been tested for change detection outside the OSCD dataset. Therefore, we designed and developed a Siamese CNN similar to that presented in \cite{daudt2018urban} and trained and tested in the OSCD dataset and obatained a change detection map for Mekelle city by feeding two high resolution images.

\section{Materials and Methods}

\subsection{Dataset}
The dataset we used in this research work is comprised of two different images:
\begin{enumerate}
\itemsep0em
\item OSCD dataset used for training our deep Siamese CNN.
\item High-resolution images of Mekelle city used as input to our trained model to generate the change map.
\end{enumerate}

The OSCD dataset is a collection of multispectral images captured by Sentinel-2 satellites of different cities around many countries around the globe that shows the expansion of urbanization these cities underwent. Sentinel satellites are specifically utilized for the Copernicus program, which is one of ESA's earth observation programs mainly working to enhance the management of the environment and climate change. Satellites launched for this program include weather satellites and satellites for land and ocean monitoring. The OSCD data is obtained from Sentinel-2 satellites designed to obtain high resolution (currently 60 m, 20 m, and 10 m) optical images for land monitoring worldwide. Some of the cities in the dataset include Agua Scalaras, Rio, Paris, Mumbai, Beirut, Hong Kong. It was first made available by \cite{daudt2018urban} for researchers and practitioners working in the field of computer vision and remote sensing and interested in tackling the problem of change detection. The dataset contains pairs of multispectral aerial images with changes manually annotated at the pixel level.

\begin{figure}[!ht]
\centering

\includegraphics[scale=.34]{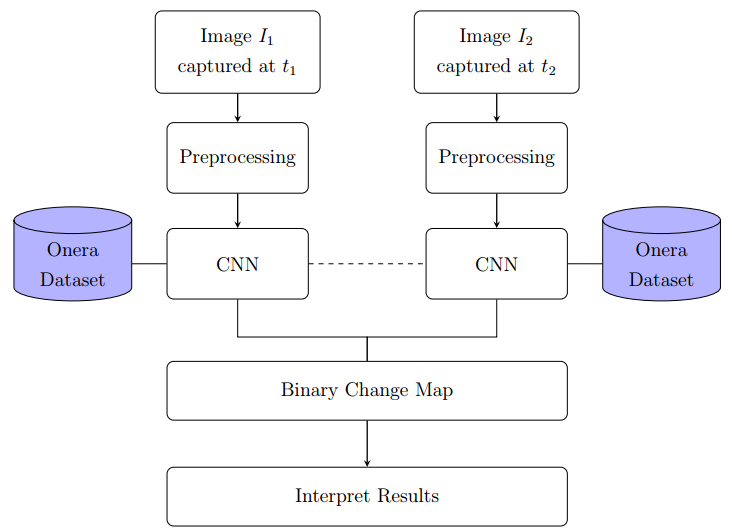}
\caption{\label{meth_fig} Methodology of urban expansion detection system.}
\end{figure}

The pairs of images are of 13-band multispectral images with spatial resolutions of 10m, 20m, and 60m. There are twenty-four regions of size 600x600 pixels at 10m resolution for each city and each of these images is registered so that it can be used directly for training. The images were captured to make sure that there was no or small amount of cloud present. The second type of data we used was a pair of images collected from Sentinel-2 satellites of the Mekelle city captured at different times. The Sentinel-2 imagery is obtained through the means of the Open Access Hub, which is part of the ESA Copernicus program. The Open Access Hub's main goal is to provide free and open access imagery data captured with different types of Sentinel satellites.

\subsection{Preprocessing}
Preprocessing of remote sensing images is an important step for building robust change detection systems. The OSCD image we used for training is already preprocessed and requires no further preprocessing; however, the Sentinel-2 image has to be preprocessed before fed into the CNN model. 

\subsubsection{Harvesting and Cropping Sentinel--2 Images}
The Sentinel-2 data images of Mekelle city were downloaded using \texttt{Sentinelsat}: an Application Programming Interface (API) built using Python. Python is an interpreted, high-level, and general-purpose programming language created by  Guido van Rossum in 1991. It supports several programming paradigms including functional programming, Object-Oriented Programming (OOP), imperative and procedural styles. Recently, the language has been widely used in different industrial and academic sectors for simulation and real-world applications. Specifically, the language has been applied in web applications, Artificial Intelligence (AI), data science, machine learning, geospatial technology, computer vision, and robotics among others. The language's success and popularity are mainly attributed to its easy to use of syntax, flat learning curve, the capability to express concepts with fewer lines of codes, and strong and matured community. To download satellite images of Mekelle city using the \texttt{sentinelsat} API, we followed the following steps.

\subsubsection*{Step1: Specify the area-of-interest (AOI)}
To download the Sentinel-2 Satellite image of Mekelle city, we first specified the boundary area of the city. We maintained a bounding polygon in geoJSON file format provided by \href{http://geojson.io/}{geojson.io}, a website that provides bounding boxes freely geospatial location in different formats such as geoJSON, Shapefile, CSV, etc. A geoJSON is a data exchange format based on JavaScript Object Notation (JSON), which contains different types of geospatial attributes. 

We considered a specific rectangular area of Mekelle that that is highly urbanized with a total area of 59.78 km$^2$. This area includes all kebelles, campuses, and most asphalts of the city. We didn't consider the rest of the areas because 1) no major urbanization has taken place in these areas for five years 2) including more areas increases the memory and mainly compute power for processing and implementing urbanization changes which we couldn't provide. 

\subsubsection*{Step 2: Download file with \texttt{sentinelsat} API}
Next, we obtained Sentinel-2 imagery automatically, through the Open Access Hub, which is part of the ESA Copernicus program, using \texttt{sentinelsat} by providing our login credentials, sensing periods, sensing modes, satellite platform, and product type. Since this is a change detection task, we specified two range dates: the first range is between May 14, 2015, and May 20, 2015, and the second between May 14, 2019, and May 20, 2019, with an exact of five-year differences. The script used for downloading is given as follows:

\subsubsection*{Step 3: Crop the image with \texttt{medusa} API}
The images downloaded using Step 2 contain a lot of data for that area with 13 different bands in different resolutions--10m, 20m, 60m. Further, the downloaded images represent areas beyond the specified AOI. We cropped the satellites images using the \texttt{medusa} toolbox, which a python package based \texttt{sentinelsat} API. The images are cropped according the geojson geometry we used in step 2 for Mekelle.

\subsubsection*{Step 4: Cloud reduction}
It is usually a good practice to have a dataset with very few or no clouds present in the images. The \texttt{sentinelsat} API allows users to control the amount of clouds present in the images. To that end, we set the percentage of clouds present to be between 0 and 20\%. We also verified manually the clouds present is not too large.

\subsubsection*{Step 5: Image coeregsiteration with \texttt{GEFolki} toolbox}
Coregistration is an important preprocessing step that improves the accuracy of change detection systems. In this work, we used the \texttt{GEFolki} toolbox to register images. \texttt{GeFolki} is a coregistration software developed in the framework of the MEDUSA project, first for SAR/SAR co-registration, then for other cases of remote sensing image coregistration, including heterogeneous image co-registration (such as LIDAR/SAR, optics/SAR, hyperspectral/optics, etc). \texttt{GeFolki} is implemented based on a local method of optical flow derived from the Lucas-Kanade algorithm, with a multi-scale implementation,
and specific filtering including rank filtering, rolling guidance
filtering and local contrast inversion \cite{brigot2016adaptation}. 

\subsection{Segmenting and classifying urbanized areas}
After the images are preprocessed, we segment and classify the Sentinel-2 image of Mekelle city into urbanized and nonurbanized areas. Here, urbanized indicates buildings and roads/asphalts and nonurbanized are the remaining areas including bare soil, grasses, crops, hills, water, etc. Sentinel-2 images contain a total of 13 bands with resolutions of 10m, 20, and 60m. Before we started segmentation and classification, we merged the bands with 10m resolution together (i.e. B2 (red), B3 (green), B4 (blue), and B8 (NIR)). We segmented the merged raster using an algorithm known as \textbf{simple linear iterative clustering (SLIC)} \cite{achanta2012slic}, which is a superpixel segmentation algorithm based on k-means and also created an object-based training and testing samples of vector shape using QGIS. We used these points as our truth data for training a supervised classifier known as random forest to classify urban and nonurban areas and test it using the remaining samples. 

\begin{figure}
    \centering
    \subfigure[Area of interest]
    {
        \includegraphics[width=1.3in]{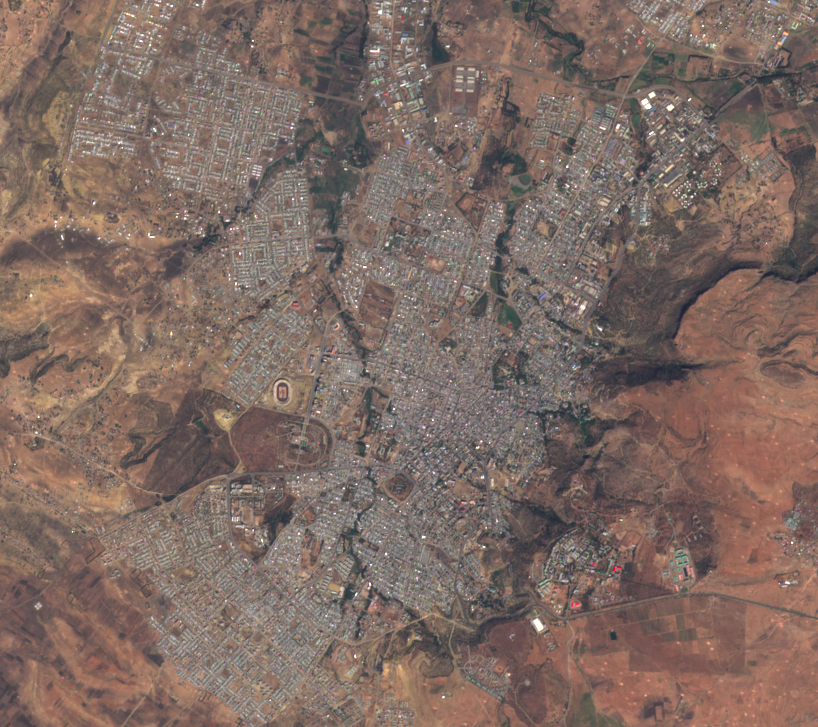}
        \label{fig:sega}
    }
    \\
    \subfigure[Training and testing samples]
    {
        \includegraphics[width=1.3in]{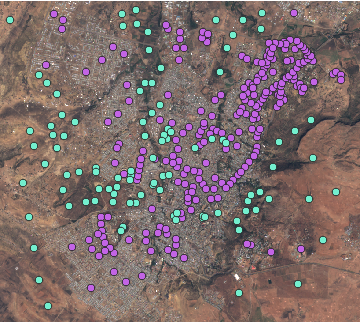}
        \label{fig:segb}
    }
    \subfigure[Segmentation results]
    {
        \includegraphics[width=1.3in]{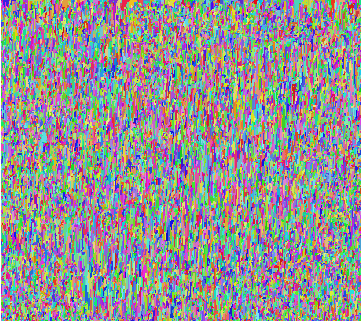}
        \label{fig:segc}
    }
    \caption{Segmentaion and classification procedure}
    \label{fig:segs}
\end{figure}

To extract features of the landcover of Mekelle city, we sampled urban and nonurban points as shown in Figure \ref{fig:segb}. We sampled 234 points representing urban objects and 101 nonurban objects. From this, 70\% were used for training the Random forest algorithm and 30\% for testing. To easily generalize for the rest of the landcover, we segmented the satellite image of Mekelle city using a SLIC algorithm that groups superpixels that share similar characteristics. The SLIC algorithms segment the input satellite images into groups of pixels that have similar perceptual meanings as shown in Figure \ref{fig:segb}. The SLIC algorithm has two hyperparameters, known as the compactness that trades off color similarity and proximity, and the number of segments, which chooses the number of centers for k-means. These two hyperparameters require careful tuning to come up with good segmentation results. With trial and error, we chose the compactness of 0.1 and the number of segments to be 750000 as shown in Figure \ref{fig:segc}.

\subsection{Training and feature extraction}

Siamese CNN architectures have been widely used in computer vision for comparing images. Recent research interests have risen to implement these architectures for change detection domains as well. Our proposed Siamese CNN model is trained and tested on the OSCD dataset and the urban expansion image is generated using two images, captured at two different times of the same scene, as input into the trained model after the images are preprocessed. The OSCD dataset is labeled at pixel level as change and no change; the changed pixels indicate changes that occurred due to artificialization only -- ignoring other changes that occur due to vegetation, disasters, etc. 

This dataset makes the best candidate to be used to train models that can be used for urban expansion applications. The Siamese network consists of two twin CNN architectures that share the same weights followed by a common fully connected layer and a softmax of two classes that outputs change or no change. The images captured at different times are fed to each network and the output layer produces changed and unchanged pixels of the scene.

The input to our Siamese network is a pair of patch images taken from the two satellite images captured at different times. Each branch of the network takes as input one of the two the patches and then applies a series of convolutional, ReLU, and max-pooling layers which are discussed next chapter. The branch outputs are concatenated together to form the binary change map of size equal to the patch size which represents the changed and unchanged pixels due to urbanization in the patches. The network classifies the patch based on artificialization changes only ignoring natural changes. This is repeated to the overall patches across the input images to output the final change map.

After the generation of change map as a binary 2D image, different evaluation measures are used to analyze the urban dynamics of the scene. The accuracy of the detection model is used to indicate the ratio of the pixels that are classified correctly, and the area of land that has been urbanized is also analyzed and reported. The architecture and training procedures are discussed in detail in the next chapter.

\section{Our Proposed Siamese CNN Model}
In chapter 2, we discuss in detail how CNN models can be used to classify and recognize patterns in images. We can use this powerful model to recognize objects and shapes in remote sensing images. Particularly, varieties of these models can be applied to recognize temporal changes in urban and gain insight into how urbanization is expanding. In this work, we propose a \textbf{Siamese CNN} to detect the change in two pairs of satellite images of Mekelle city captured at different times. A siamese CNN model contains two twin CNNs each presented with an and go through two different sets of layers that share the same weights.

\begin{figure*}[h]
\centering
  \includegraphics[width=.8\textwidth, height=6cm]{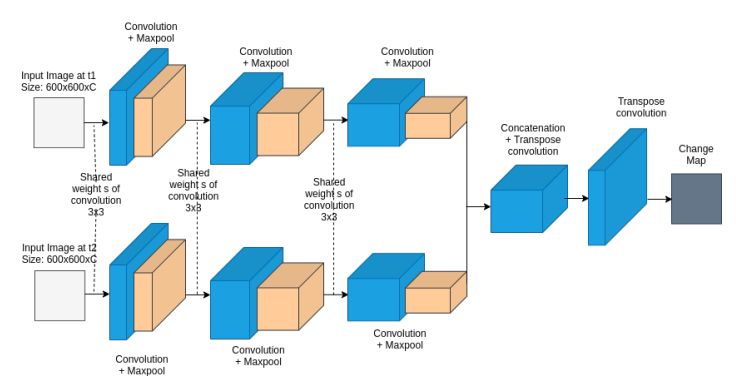}
\caption{\label{proposed} Architecture of the proposed model for change detection. Note blue colors indicate convolution, yellow colors indicate max-pooling}
\end{figure*}

The siamese CNN is mainly used for feature extraction. since CNN takes one image patch as input and produces some encoded representation of that patch, the siamese CNN extracts features from both pairs of the same scenes taken at different times. Figure \ref{proposed} shows our proposed model and it is inspired by the works of \cite{daudt2018urban}. But to make it more suitable to our objective we made some modifications and optimizations to the original architecture. 

Each branch in the networks is fed one of the patches as input. The patches small subset pixels from the satellite image that are small enough to represent a single object such as building, road, etc. Since the two branches of the network share the same weight, they extract features from the two patches using the same approach. This is natural to our image since we are using optical images that are homogenous: captured by the same type of sensor and have similar characteristics.

The network is composed of two stages: in the first stage, the input goes through a constant increase in depth and decreases in size using convolutional directly followed by max-pooling operations. At each layer, we used the $3 \times 3$ kernels for convolution with padding of size 1; the convolutional has the effect of simply increasing the depth (number of channels) of the input at each layer. Then this is followed by a max-pool of kernel size $2 \times 2$ with a stride of size 2, which has the effect of reducing the input size by half. In the second stage, we have operations that take the difference between the two final outputs of the first stage and perform transpose operation to restore the original size of the input. The difference of the input volumes is based on the euclidean distance of the pixels and is used to show if changes have occurred or not in that particular pixel. If the distance is very large it means there a high probability that change has occurred in the pixels; otherwise, no change has occurred. The final output change map is obtained by implementing a log softmax operation on the final convoluted volume. The change map has either 0 (no change) or 1 (for change) pixel values.

\section{Discussion}

\subsection{Landcover Classification Results}

\begin{figure}[!ht] 
\centering
\includegraphics[width=0.3\textwidth]{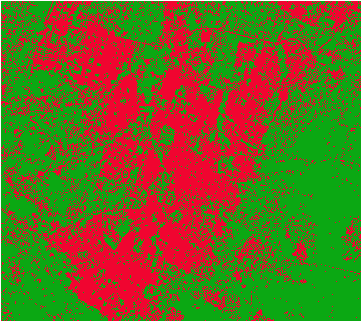}%
\caption{\label{seg} Landcover classification results.}
\end{figure}

When the classifier algorithm is trained and tested, it can easily generalize to each segment by mapping a point that belongs in one segment to the rest of the points as belonging to the same class. The final classification of the landcover as urbanized/nonurbanized is shown in Figure \ref{seg}. 

\begin{figure}[!ht] 
\centering
\includegraphics[width=0.45\textwidth]{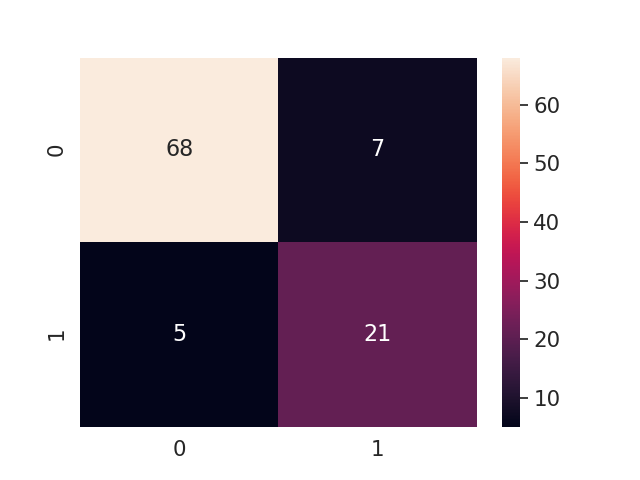}%
\caption{\label{cm} The confusion matrix of Landcover classifier.}       
\end{figure}

The performance of the landcover classifier is depicted in Figure \ref{cm} as a confusion matrix. 0 class labels represent urban points and 1 for nonurban points. The testing results show that out of 75 total urban vector points, 68 were accurately classified and 7 misclassified as nonurban. And out of 26 nonurban points, 21 were correctly classified and 5 misclassified as urban. This results in an accuracy of 93.15\%.

\subsubsection{Training and Testing Accuracy and Loss of Siamese Network}

\begin{figure}
    \centering
    \subfigure[Epoch vs Accuracy]
    {
        \includegraphics[width=2in]{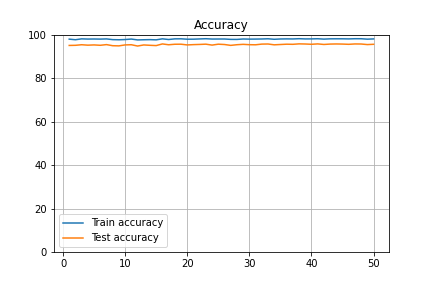}
        \label{fig:acc1}
    }\hfill
    \subfigure[Epoch vs Loss]
    {
        \includegraphics[width=2in]{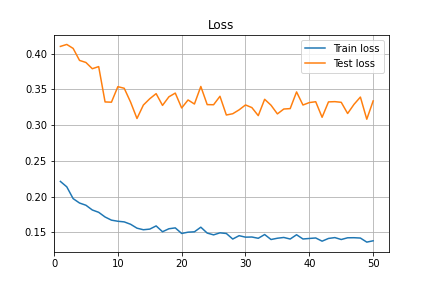}
        \label{fig:acc2}
    }
    \caption{\label{accloss} The relationship of epoch and accuracy, and epoch and loss during traing.}
    \label{fig:segs}
\end{figure}

The performance of our model during training and testing is depicted in Figure \ref{accloss}. Figure \ref{fig:acc1} shows the accuracy is slightly increasing during training and testing across the epochs. More specifically, the model scored 98.5\% accuracy during training and 97.2\% accuracy during testing. This indicates that the overall performance of the model is good and is not affected by overfitting during training since the accuracy is slowly increasing during training and testing. 

Furthermore, Figure \ref{fig:acc2} shows the performance of the model in terms of loss in which case there is an overall decrease in loss both during training and testing. The loss of the final model is slightly less during training than testing (0.14 vs 0.33). The gap is however small and is an indication that the model is not suffering from overfitting since the losses during both phases are decreasing at a similar rate across the epochs. In general, the model shows that it has an overall good performance in terms of loss and accuracy at the training and testing phase.

\subsection{Evaluation of Change Detection for Mekelle City}

\begin{figure}
    \centering
    \subfigure[Mekell City taken at May 10, 2015]
    {
        \includegraphics[width=1.3in]{pair1}
        \label{fig:c1}
    }
    \subfigure[Mekelle City taken at May 10, 2019]
    {
        \includegraphics[width=1.3in]{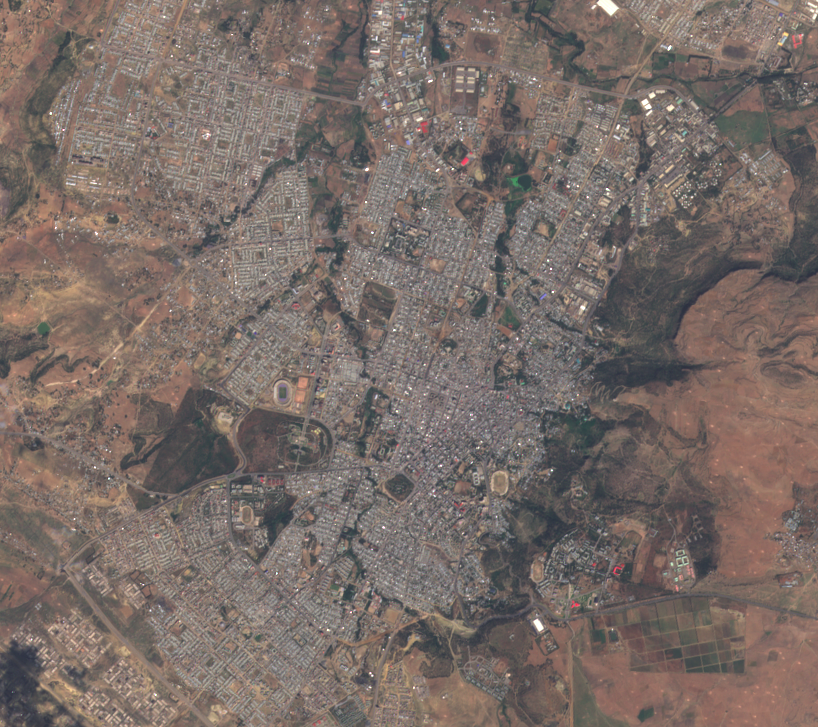}
        \label{fig:c2}
    }
    \subfigure[Change Map]
    {
        \includegraphics[width=1.3in]{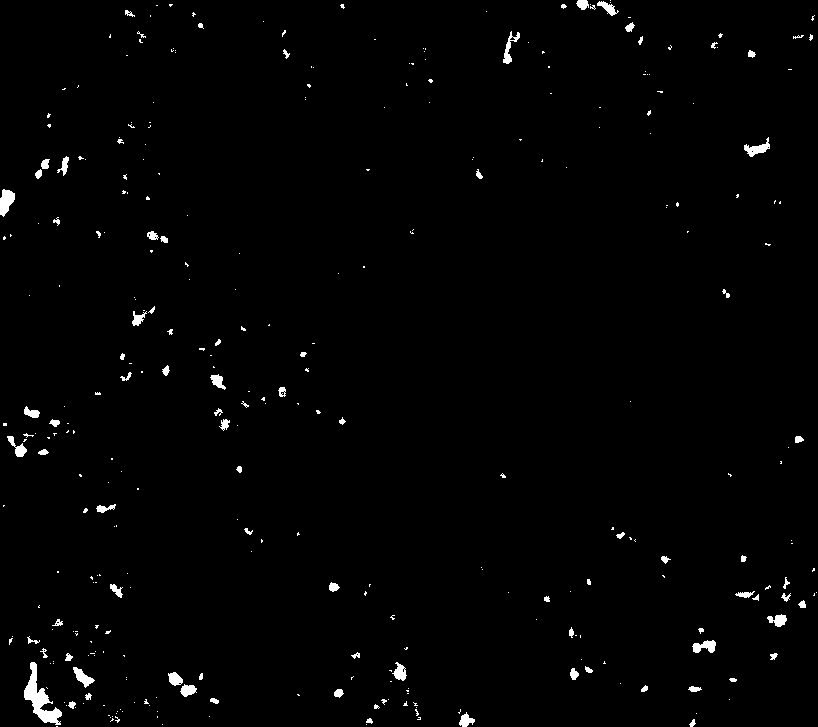}
        \label{fig:c3}
    }
    \caption{\label{change} The inputs of Mekelle city taken at different times and the resulting change map depicting the changes due to urbanization}
    
\end{figure}

The final trained change detection model was fed to Sentinel-2 satellite images of Mekelle city taken in May 2015, and May 2019. As shown in Figure \ref{change}, when the two temporal images (Figure \ref{fig:c1} and Figure \ref{fig:c2}) are fed into the network, the network outputs a change map show in Figure \ref{fig:c3}. The white pixels indicate changes between the two dates and the black changes indicate no changes due to urbanization. 

\subsection{Quantitative Change Detection of Mekelle City}

\begin{figure}[h] 
\centering
\includegraphics[width=0.75\linewidth]{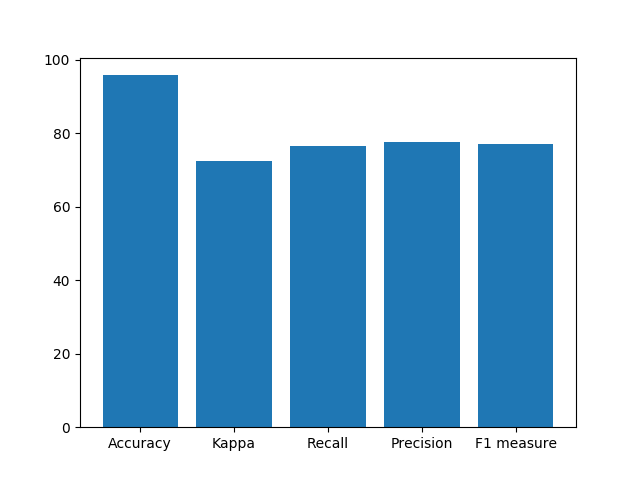}%
\caption{\label{perfor} Summary of Mekelle city Urban change detection performance}
\end{figure}

Figure \ref{perfor} shows the overall performance of the urban changes for five years in terms of Overall accuracy, Kappa measure, Recall, Precision, and F1 measure. As shown in the figure the model scored 95.8\% overall accuracy which is similar to the accuracy scored during training and testing. Moreover, the Kappa measure is 72.5\%. The Kappa value measures agreement between the actual and the observed values. A value of 72.5\% is in general considered a good agreement. The recall value which is an indicator of how much unchanged pixels are classified as changed is 76.5\% a low value in general indicating the probability of unchanged pixels classified as changed is high; in our case, we achieved a more than average value of recall. The precision score is 77.7\%; an indicator of how much changed pixels are classified as unchanged. The precision value is slightly higher than the recall which suggests that unchanged pixels are more likely to be classified as changed than the other way around. Finally, we have the F1 measure which is 77.1\%. The F1 measure is a good measure when we unbalanced number of classes, which is true in our case. The score is slightly above average indicating a generally good detection capability of the model. Finally, we have the F1 measure which is 77.1\%. The F1 measure is a good measure when we unbalanced number of classes, which is true in our case. The score is slightly above average indicating a generally good detection capability of the model.

The overall change detection that happend in the span of the span of five years can be analyzed from Figure \ref{change}. A total area of 736891.2m$^2$ was urbanized during these years which accounts for 1.23\% of the total area.

\bibliographystyle{IEEEtran}
\bibliography{cdcnn}

\end{document}